%% file: main.tex
\documentclass[11pt,a4paper]{article}
\usepackage{authblk}
\usepackage{emnlp2018}
\aclfinalcopy 
\usepackage{times}
\usepackage{latexsym}
\usepackage{xr}

\usepackage{url}
\usepackage{thanks-nostar}

\usepackage{multirow}
\usepackage{booktabs}

\usepackage[utf8]{inputenc}

\usepackage[disable]{todonotes} 

\newcommand{\note}[4][]{\todo[author=#2,color=#3,size=\scriptsize,fancyline,#1]{#4}} 
\newcommand{\huda}[2][]{\note[#1]{Huda}{violet!40}{#2}}
\newcommand{\Huda}[2][]{\huda[inline,#1]{#2}}
\newcommand{\xuan}[2][]{\note[#1]{Xuan}{cyan!40}{#2}}

\newcommand{\marine}[2][]{\note[#1]{Marine}{yellow!40}{#2}}
\newcommand{\Marine}[2][]{\marine[inline,#1]{#2}}
\newcommand{\gaurav}[2][]{\note[#1]{Gaurav}{blue!40}{#2}}
\newcommand{\Gaurav}[2][]{\gaurav[inline,#1]{#2}}

\newcommand{\marianna}[2][]{\note[#1]{Marianna}{red!40}{#2}}
\newcommand{\Marianna}[2][]{\marianna[inline,#1]{#2}}

\title{An Empirical Exploration of Curriculum Learning \\for Neural Machine Translation}

\author[1]{{Xuan Zhang\thanks{\ \ Corresponding author. \texttt{xuanzhang@jhu.edu}}} }
\author[1]{Gaurav Kumar}
\author[1]{Huda Khayrallah}
\author[2]{Kenton Murray}
\author[3]{Jeremy Gwinnup}
\author[4]{\\Marianna J Martindale}
\author[1]{Paul McNamee}
\author[1]{Kevin Duh}
\author[4]{Marine Carpuat}
\vspace{-10pt}
\affil[1]{Johns Hopkins University}
\affil[2]{University of Notre Dame}
\affil[3]{Air Force Research Laboratory}
\affil[4]{University of Maryland}

\date{}

\begin{document}
\maketitle
\begin{abstract} 
Machine translation systems based on deep neural networks are expensive to train. Curriculum learning aims to address this issue by choosing the order in which samples are presented during training to help train better models faster. We adopt a probabilistic view of curriculum learning, which lets us flexibly evaluate the impact of curricula design, and perform an extensive exploration on a German-English translation task. Results show that it is possible to improve convergence time at no loss in translation quality. However, results are highly sensitive to the choice of sample difficulty criteria, curriculum schedule and other hyperparameters. 
\end{abstract}

\input{introduction}
\input{background}

\input{probview}
\input{criteria}
\input{methods}

\input{setup}

\input{results}


\bibliography{references,clreferences}
\bibliographystyle{acl_natbib_nourl}
\clearpage
\onecolumn
\appendix
\input{sm}

\end{document}

%% file: introduction.tex
\section{Introduction}
\label{sec:introduction}

Curriculum learning  \citep{BengioLouradourCollobertWeston2009} hypothesizes that choosing the order in which training samples are presented to a learning system can help train better models faster. In particular, presenting samples that are easier to learn from before presenting difficult samples is an intuitively attractive idea, which has been applied in various ways in Machine Learning and Natural Language Processing tasks \citep[inter alia]{BengioLouradourCollobertWeston2009,TsvetkovFaruquiLingMacWhinneyDyer2016,CirikHovyMorency2016,GravesBellemareMenickMunosKavukcuoglu2017}. 

In this paper, we conduct an empirical exploration of curriculum learning for Neural Machine Translation (NMT). NMT is a good test case for curriculum learning as training is prohibitively slow in the large data conditions required to reach good performance \citep{KoehnKnowles2017}. However, designing a curriculum for NMT training is a complex problem. First, it is not clear how to quantify sample difficulty for this task. Second, NMT systems already rely on established data organization methods to deal with the scale and varying length of training samples \citep{KhomenkoShyshkovRadyvonenkoBokhan2016,doetsch2017comprehensive,SennrichFiratChoBirchHaddowHitschlerJunczys-DowmuntLaubliMiceliBaroneMokryNadejde2017,hieber2017sockeye}, and it is not clear how a curriculum should interact with these existing design decisions.
\citet{KocmiBojar2017} showed that constructing and ordering mini-batches based on sample length or word frequency helps when training for one epoch. It remains to be seen how curricula impact training until convergence.

To address these issues, we adopt a probabilistic view of curriculum learning that lets us explore a wide range of curricula flexibly. Our approach does not order samples in a deterministic fashion. Instead, each sample has a probability of being selected for training, and this probability changes depending on the difficulty of the sample and on the curriculum's schedule. We explore difficulty criteria based on NMT model scores as well as linguistic properties. We consider a wide range of schedules, based not only on the easy-to-difficult ordering, but also on strategies developed independently from curriculum learning, such as dynamic sampling and boosting \citep{ZhangKimCregoSenellart2017,WeesBisazzaMonz2017,WangUtiyamaSumita2018a}.


We conduct an extensive empirical exploration of curriculum learning on a German-English translation task, implementing all training strategies in the Sockeye NMT toolkit.\footnote{Sockeye is a state-of-the-art open-source NMT framework at \url{https://github.com/awslabs/sockeye} Our modification is publicly available at \url{https://github.com/kevinduh/sockeye-recipes/tree/master/egs/curriculum}}. Our experiments confirm that curriculum learning can improve convergence speed without loss of translation quality, and show that viewing curriculum learning more flexibly than strictly training on easy samples first has some benefits.  We also demonstrate that curriculum learning is highly sensitive to hyperparameters, and no clear single best strategy emerges from the experiments.

In this sense, our conclusions are both positive and negative: We have confirmed that curriculum learning can be an effective method for training expensive models like those in NMT, but careful design of the specific curriculum hyperparameters is important in practice. 

%% file: background.tex
\section{Related Work}
\label{sec:background}

\Marine{Does this make more sense just after intro as motivating background or at the end as related work?}
\Huda{my thought would be to put it at the start (leave it where it is) but have it titled related work?}
\Marine{terminology: we'll talk about easy vs difficult samples. Avoid  ``complexity'' and ``hardness''}
\Gaurav{I think, we should use difficulty instead of complexity/hardness. Hence, change "organizing training samples based on complexity" to "organizing training samples based on difficulty". I am making this modification but marking it with the tag textup. If you approve, remove this tag.}
\citet{BengioLouradourCollobertWeston2009} coined the term of curriculum learning to refer to techniques that guide the training of learning systems ``by choosing which examples to present and in which order to present them in the learning system'', and hypothesize that training on easier samples first is beneficial. While organizing training samples based on \textup{difficulty} has been demonstrated in NLP outside of neural models \---\ e.g., \citet{SpitkovskyAlshawiJurafsky2010} bootstrap unsupervised dependency parsers by learning from incrementally longer sentences \---\, curriculum learning has gained popularity to address the difficult optimization problem of training deep neural models \citep{Bengio2012}. \citet{BengioLouradourCollobertWeston2009} improve neural language model training using a curriculum based on increasing vocabulary size. More recently, \citet{TsvetkovFaruquiLingMacWhinneyDyer2016} improve word embedding training using Bayesian optimization to order paragraphs in the training corpus based on a range of distributional and linguistic features (diversity, simplicity, prototypicality). 

While curriculum learning often refers to organizing examples from simple to \textup{difficult}, other data ordering strategies have also shown to be beneficial: \citet{AmiriMillerSavova2017} improve the convergence speed of neural models using spaced repetition, a technique inspired by psychology findings that human learners can learn efficiently and effectively by increasing intervals of time between reviews of previously seen materials.

Curriculum design is also a concern when deciding how to schedule learning from samples of different tasks either in a sequence from simpler to more \textup{difficult} tasks \citep{CollobertWeston2008} or in a multi-task learning framework \citep{GravesBellemareMenickMunosKavukcuoglu2017,KiperwasserBallesteros2018}. In this work, we focus on the question of organizing training samples for a single task.

In NMT, curriculum learning has not yet been explored systematically. In practice, training protocols randomize the order of sentence pairs in the training corpus  \citep{SennrichFiratChoBirchHaddowHitschlerJunczys-DowmuntLaubliMiceliBaroneMokryNadejde2017,hieber2017sockeye}. 
There are works that speed training up by batching the samples of similar lengths \citep{KhomenkoShyshkovRadyvonenkoBokhan2016,doetsch2017comprehensive}.
Such works attempt to improve the {\em computational efficiency}, while curriculum learning is supposed to improve the {\em statistical efficiency} --- fewer batches of training examples are needed to achieve a given performance. 

\citet{KocmiBojar2017} conducted the first study of curriculum learning for NMT by exploring the impact of several criteria for curriculum design on the training of a Czech-English NMT system for one epoch. They ensure samples within each mini-batch have similar linguistic properties, and order mini-batches based on complexity. They show translation quality can be improved by presenting samples from easy to hard  based on sentence length and vocabulary frequency. However, it remains to be seen whether these findings hold when training until convergence.

Previous work has focused on dynamic sampling strategies, emphasizing training on samples that are expected to be most useful based on model scores or domain relevance.
Inspired by boosting \citep{Schapire2002},  \citet{ZhangKimCregoSenellart2017}, at each epoch, assign higher weights to training examples that have lower perplexities under the model of previous epoch.
Similarly, \citet{WeesBisazzaMonz2017} and \citet{WangUtiyamaSumita2018a} improve the training efficiency of NMT by dynamically select different subsets of training data between different epochs. The former performs this dynamic data selection according to domain relevance \citep{AxelrodDataSelection} while the latter uses the difference between the training costs of two iterations.

Taken together, these prior works show that sample difficulty can impact NMT, but it remains unclear how to balance the benefits of existing sample randomization and bucketing strategies with intuitions about sample ordering, as well as which ranking criteria and strategies should be used. We revisit these ideas in a unified framework, via experiments on a German-English task, training until convergence.

%% file: probview.tex
\section{A Probabilistic View of Curriculum Learning}\label{sec:probview}

Let $(x,y)$ be a bitext example, where $x$ is the source sentence and $y$ is the target reference translation.  
We use subscripts $i$ to denote the sample index and assume a training set $\mathcal{D} = \{(x_i,y_i)\}_{i=1,2,\ldots S}$ of size $S$. 
Curriculum learning can be formulated in a probabilistic view, where each sentence pair $(x_i,y_i)$ has a probability of being selected for training, and this sampling probability changes depending on the \textup{difficulty} of the example and the curriculum schedule  \citep{BengioLouradourCollobertWeston2009}. 

\begin{figure}[ht]
	\centering
	\includegraphics[width=\linewidth]{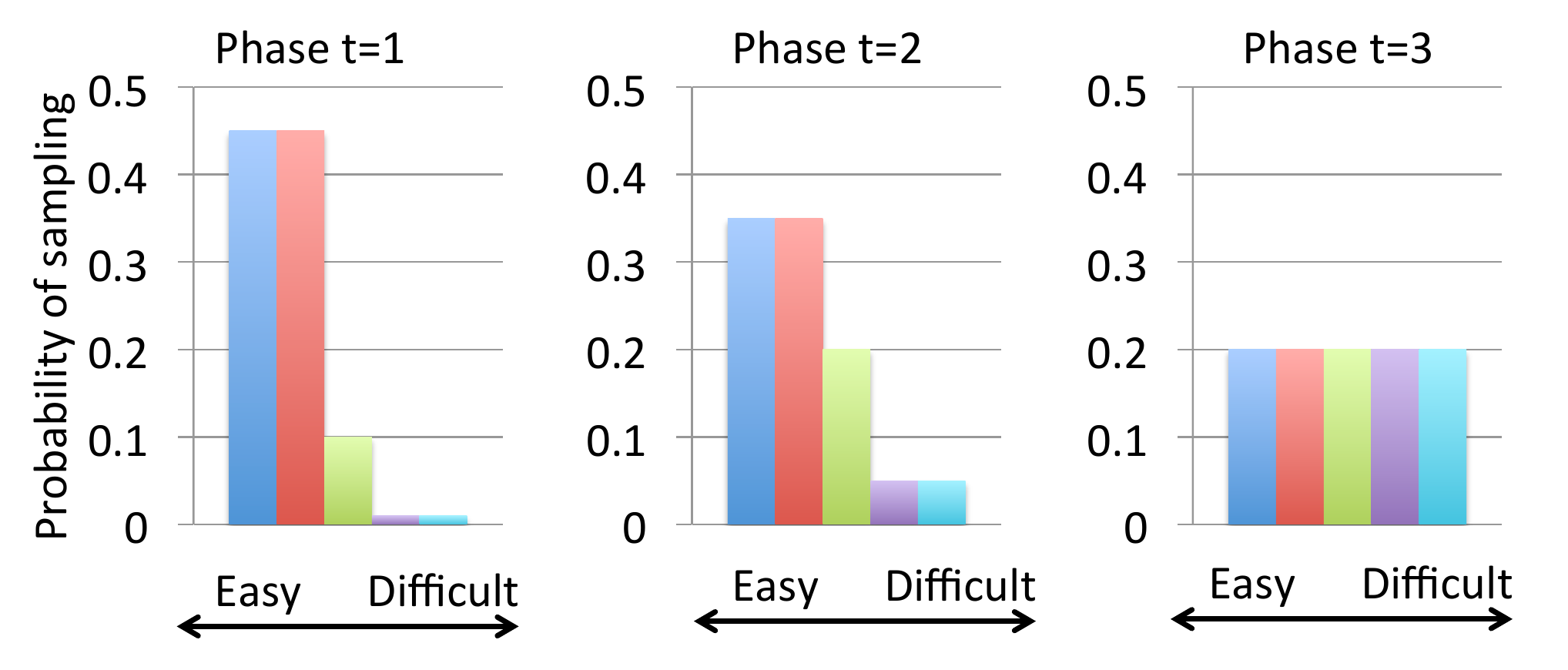}
    \caption{Probabilistic view of curriculum learning: On the x-axis, the examples are arranged from easy to difficult. y-axis is the probability of sampling the example for training. By specifying different kinds of sampling distributions at different phases, we can design different curriculums. In this example, $t=1$ samples from the first three examples, $t=2$ includes the remaining two examples but at lower probability, and $t=3$ defaults to uniform sampling (regardless of difficulty).}
\label{fig:probabilistic_curriculum}
\end{figure}

Specifically, we segment the curriculum schedule into distinct phases $t$ which correspond to different time points during training. 
For instance, $t=1$ could be the first $N$ checkpoints, $t=2$ is the next $N$ checkpoints, etc. 
The definition of phases is flexible: alternatively $t=1$ may correspond to the first epoch, and $t=2$ may correspond to the second epoch (or more). 
At each phase $t$, we maintain a multinomial distribution $q_i^t$ over the examples in $\mathcal{D}$, 
i.e.~ $\sum_{i=1}^S q_i^t = 1$ and $q_i^t \geq 0~\forall i$.
To implement the curriculum schedule that begins with easy examples, we would start at $t=1$ by setting $q_i^t$ to be high for easy examples and $q_i^t$ to be low (or zero) for difficult examples. 
Gradually, for large $t$, we increase $q_i^t$ for the more difficult examples. 
At some point, all examples have equal probability of being selected; this corresponds to the standard training procedure. 
An illustration of this probabilistic view of curriculum learning is shown in Figure \ref{fig:probabilistic_curriculum}. 

There are two advantages to this probabilistic sampling view of curriculum learning:
\begin{enumerate}
\item It is a flexible framework that enables the design of various kinds of curriculum schedules. By specifying different kinds of distributions, one can perform easy-to-difficult training or the reverse difficult-to-easy training. One can default to uniform sampling, which corresponds to standard training with random mini-batches. Many of these variants are described in Section \ref{ssec:curriculumschedule}. 
\item It is simple to implement in existing deep learning frameworks, requiring only a modification of the data sampling procedure. In particular, it is modular with respect to the optimizer's learning rate schedule and mini-batch shuffling mechanism; these represent best practice in deep learning, and may be suboptimal if modified. Further, the optimizer only needs access to sampling probability $q_i^t$, which abstracts away from the various difficulty criteria such as sentence length and vocabulary frequency (to be described in Section \ref{sec:rankingcriteria}). This enables us to plug-in and experiment with many criteria.
\end{enumerate}

\begin{figure}[ht]
	\centering
	\includegraphics[width=\linewidth]{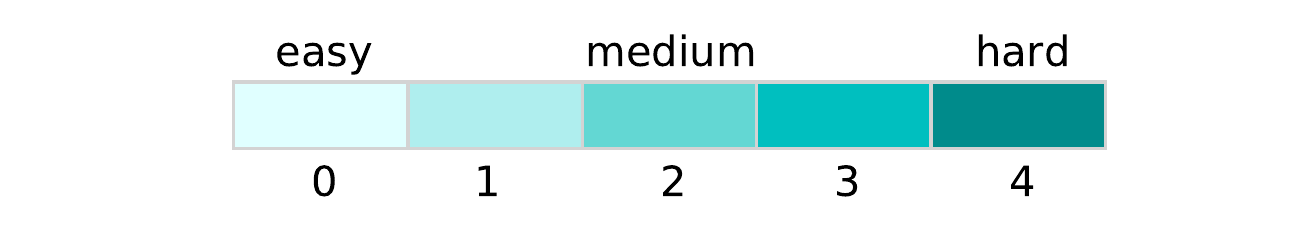}
    \caption{Training data organized by level of difficulty. Each block is a shard (i.e., a subset of the dataset) and darker shades indicate increasing difficulty. Note that the width of each patch does {\em not} indicate the number of samples in that shard, as it may vary for different difficulty criteria.}
    \label{fig:shards}
\end{figure}

Without loss of generality, in practice we recommend grouping examples into shards (Figure \ref{fig:shards}) such that those in the same shard have similar difficulty criteria values.\footnote{Shards are not to be confused with buckets (grouping of similar-length samples). Shards are simply subsets of the training data and may allow for bucketing by length within themselves.} Then we define the sampling distributions over shards rather than examples. Since there are fewer shards than examples (e.g., 5 shards vs. 1 million examples for a typical-sized dataset), the distributions are simple to design and visualize. Sharding is described in more detail in Section \ref{ssec:datareorganization}.

\begin{figure*}[h]
	\centering
    \includegraphics[width=\linewidth]{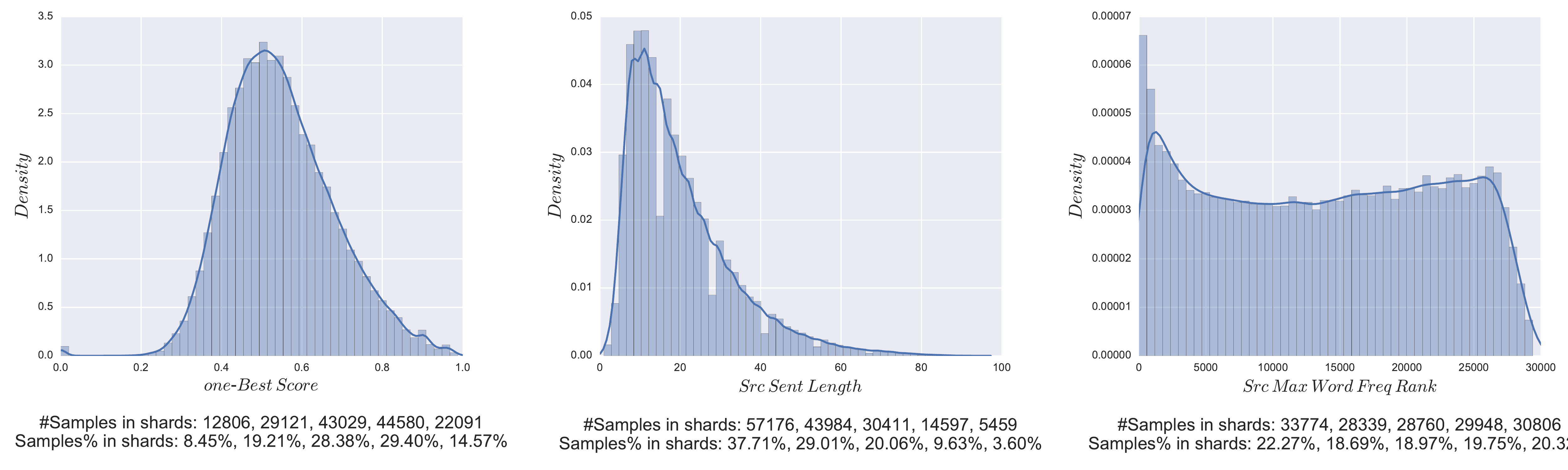}
    \caption{Difficulty score distribution on DE-EN TED Talks training set (151,627 sentence pairs in total) scored by selected difficulty criteria. Sharding results generated from Jenks Natural Breaks classification algorithm are shown below each subplot, in the ascending order of difficulty levels. }
	\label{fig:datastat3}
\end{figure*}

%% file: criteria.tex
\section{Sample Difficulty Criteria}\label{sec:rankingcriteria}
In this work, we quantify the translation difficulty of a sentence pair by two kinds of criteria (or score\footnote{Criteria and score are interchangeable in this paper.}): 
1) how well an auxiliary translation model captures the pair and 2) linguistic features which are orthogonal to any translation model. 

\paragraph{Model-based Difficulty Criteria} We use the {\em one-best score}, which is the probability of the one-best translation (the product of its word prediction probabilities) from an auxiliary (possibly simpler) translation model, given a source sentence. This represents $p(\hat{y}\mid x)$, where $x$ is the source sentence and $\hat{y}$ is the one-best translation. A high \textit{one-best score} for a translation suggests the auxiliary model is very certain of its prediction with small chance of choosing other candidates. Although the prediction might not be the ``correct answer", $p(\hat{y}\mid x)$ shows the confidence of the model for that prediction, and indicates how easy the prediction is according to the model.

\paragraph{Linguistic Difficulty Criteria}  Linguistic features, including sentence length and vocabulary frequency, can also be used to measure the difficulty of translating a sample \citep{KocmiBojar2017}. Short sentences usually do not have difficult syntactic structures, while lengthier sentences with long-distance dependencies are difficult to handle for NMT models \citep{hasler2017source}. To capture this phenomenon, we rank samples by the length of source and target sentence and by the sum of the length of each sentence in the pair.

\citet{sutskever2014sequence} shows that a NMT model's performance decreases on sentences with more rare words. Similar to \citet{KocmiBojar2017}, we first sort words by their frequency to get the word frequency rank, then order sentences based on the rank of the least frequent word in the sentence ({\em max word frequency rank}). Organizing sentences by this criterion is equivalent to gradually increasing the vocabulary size and training on sentences that only contain words in the current partial vocabulary \citep{BengioLouradourCollobertWeston2009}. In addition to maximizing, we also experimented with the {\em average word frequency rank}. Again, we collect word frequency rank scores for source sentences, target sentences and concatenations of both\footnote{In the concatenation, the word rank is obtained based on whether the word belongs in the source or the target; i.e., we maintain separate word frequency lists for each language.}. 

%% file: methods.tex
\section{Methods}
\label{sec:methods}
\Marine{Anyone has a better idea than ``Method'' for this section title?}

Having defined criteria for measuring sample difficulty and illustrated how they can be used in a probabilistic curriculum learning framework, we now describe in more detail how this framework was instantiated for our study. We present our approach for organizing data into shards given sample difficulty scores (Section~\ref{ssec:datareorganization}), how the shards are used by the curriculum schedule (Section~\ref{ssec:curriculumschedule}), and how this fits in the overall training strategy (Section~\ref{ssec:trainingstrategy}).


\subsection{Data Sharding} \label{ssec:datareorganization}

As described in section~\ref{sec:probview}, samples are grouped into shards of similar difficulty (Figure \ref{fig:shards}). This can be done by various methods. One approach is to set thresholds on the difficulty score \citep{KocmiBojar2017}. An alternative is to distribute the data evenly such that each shard will have same number of samples. The first approach makes it difficult to choose reasonable breaks while trying to ensure that each shard has roughly the same number of samples (Figure \ref{fig:datastat3}). In contrast, the latter may result in unwanted fluctuations in difficulty within the same shard, and not enough difference between different shards. 

We instead use the Jenks Natural Breaks classification algorithm \citep{Jenks1997}, an algorithm commonly used in Geographic Information Systems (GIS) applications \citep{brewer2006basic,chrysochoou2012gis}. This method seeks to minimize the variance within classes and maximize the variance between classes. Figure \ref{fig:datastat3} shows examples of the univariate classification results using Jenks algorithm on our training corpus (TED Talks, \citet{duh18multitarget}) where training samples are reorganized by various criteria representing difficulty (Section \ref{sec:rankingcriteria}). Distributions obtained for other complexity criteria are available in the supplementary material.

\subsection{Curriculum Schedule}\label{ssec:curriculumschedule}
\begin{figure}[ht]
	\centering
	\includegraphics[width=\linewidth]	{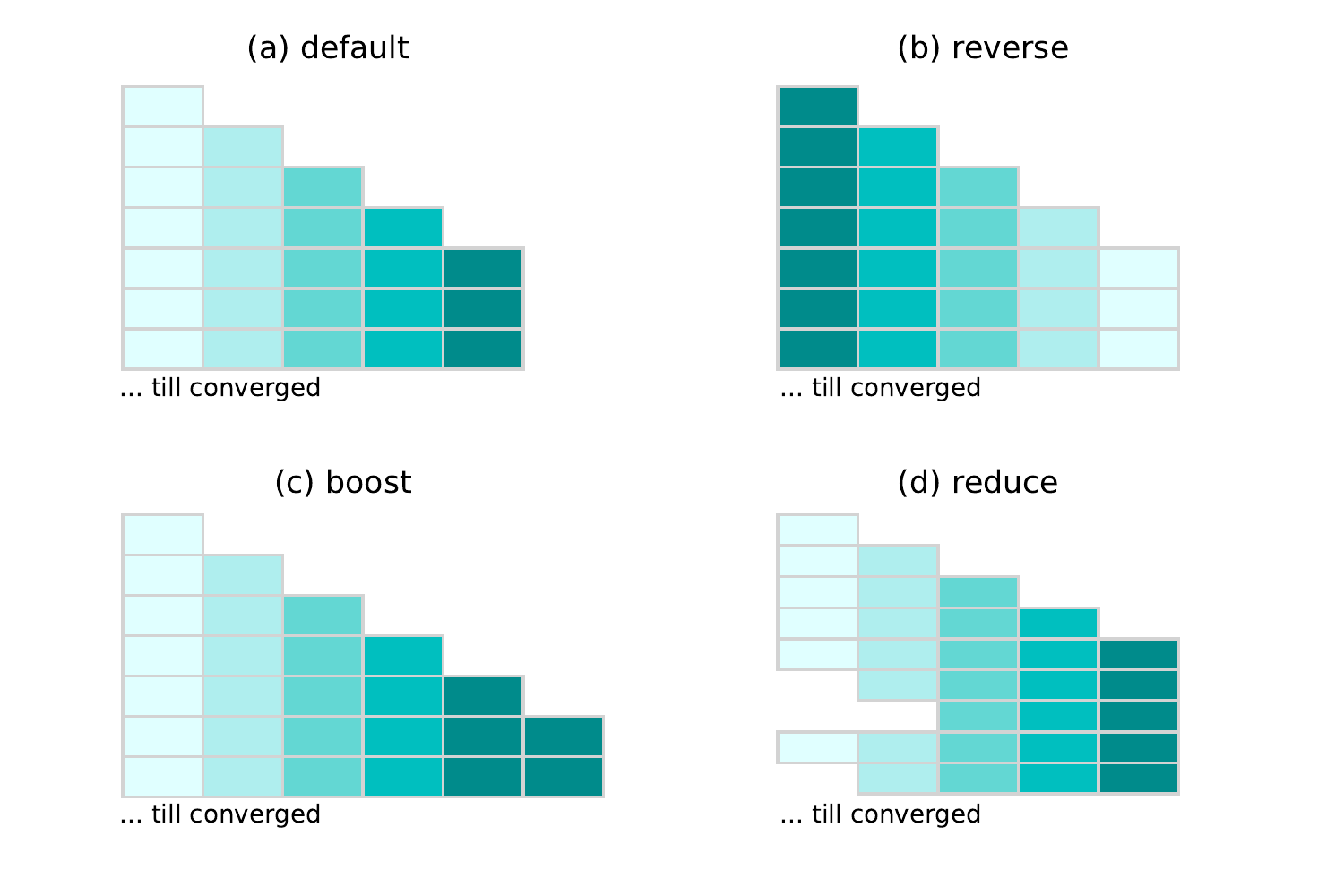}
	\caption{Training with different curriculum schedules. The colored blocks are shards of different difficulty levels (see figure \ref{fig:shards}). Within a sub-figure, each row represents a phase, and shards in that row are accessible shards based on the curriculum. Training starts from the first row and goes through the following rows in succession. Hence, at each phase only subsets of the training data and certain difficulty classes are available. Note that shards (and the samples within them) are shuffled as described in Section \ref{ssec:trainingstrategy}.  }
    \label{fig:curriculumschedule}
\end{figure}

The curriculum's {\em schedule} defines the order in which samples of different difficulty classes are presented to the learning system. A curriculum's {\em phase} is the period between two curriculum updates.\footnote{This is similar to the concept of an epoch except that only a subset of the training data may be available based on the curriculum's schedule.} For NMT models, it is natural to come up with the idea of first presenting easy samples to the models, as suggested by \citet{BengioLouradourCollobertWeston2009}. In the following sections, we refer to this as the {\em default} schedule. We also introduce four variants of the {\em default} schedule (Figure~\ref{fig:curriculumschedule}) which lets us explore different trade-offs.

\begin{itemize}
\item {\bf default} Shards are sorted by increasing level of difficulty. Training begins with the easiest shard and harder shards will be included in subsequent phases.
\item {\bf reverse} Shards are sorted in descending order of difficulty. Training begins with the hardest shard and easier shards will be included in subsequence phases.  
\item {\bf boost} A copy of the hardest shard is added to the training set, after the model has processed shards of all difficulty classes.
\item {\bf reduce} Once all shards have been visited, we start removing shards from training one at the end of each phase, starting with the easiest. Once a fixed number of shards have been removed (2 in our case), we add them back. This {\em reduce and add-back} procedure will be iteratively continued until the training converges. The effect is that the model gets to look at harder shards more often. 

\item {\bf noshuffle} Same as {\em default} except that shards are never shuffled; that is, they are always presented to the model in ascending order of difficulty (Samples within shards are shuffled as usual). 
\end{itemize}


The {\em reverse} schedule tests the assumption that presenting easy examples first helps learning. It remains unclear if we should start with the easier sentences and move to more difficult ones, or if perhaps some of the difficult sentences are too hard for the model to learn and we should focus on straightforward sentences at the end. In addition,  we are unsure of what the model will find more easy or difficult.

Another open question is whether presenting shards randomly during each curriculum phase (as done in the {\em default} schedule) weakens the curriculum. We explore an alternative by forcing the shard visiting order to be deterministic --- always starting from the easiest shard, ending at the hardest shard for this phase. We label this schedule as {\em noshuffle}, since shuffling does not occur. {\em Noshuffle} may be helpful in the sense that every time the model is assigned with a new harder shard, it will review old shards in a more organized way. This method can be viewed as restarting the curriculum at each phase.

The last two schedules are adapted from   \citet{ZhangKimCregoSenellart2017}, who improve NMT convergence speed by duplicating samples considered difficult based on model scores. The {\em boost} schedule combines the idea of training on easy samples first (from {\em default}), while putting more emphasis on difficult samples  (as in {\em reverse}).  The {\em reduce} schedule additionally makes sure that the model gets to look at difficult shards more often. This is accomplished by removing easy shards from epochs and then adding them back again later. 


\subsection{Training Strategy}\label{ssec:trainingstrategy}

Finally, we address the question of how to draw mini-batches from the training data which has been sharded based on difficulty.
Current state-of-the-art NMT model implementations bucket the training samples based on source and target length. Mini-batches are then drawn from these buckets, which are shuffled at each epoch.
One way of drawing mini-batches while conditioning on difficulty is to sort the training samples by difficulty and to then draw these deterministically starting from the easiest to the most difficult sample. However, this loses the benefits gained by shuffling the data at each epoch.


Instead, our work uses a strategy similar to the work of \citet{BengioLouradourCollobertWeston2009}. We organize samples into shards\footnote{5 shards in our experiments.} according to the univariate classification results (Section \ref{ssec:datareorganization}) and allow further bucketing by sentence length within each shard. Samples within each shard are shuffled at each epoch, ensuring that we draw random mini-batches of the same difficulty.

Given shards of different difficulty levels, we follow these steps for training:
\begin{itemize}
\item The curriculum's schedule defines which shards are available for training. We call these the {\em visible} shards for this phase of curriculum training.
\item These shards are then shuffled (except when we use the {\em noshuffle} schedule)\footnote{In shuffling, we ensure that the first shard for this phase is not the same as the last shard from the last phase.} so that the model is trained using random levels of difficulty (in contrast to always using easy to hard).
\item The samples within each shard are shuffled and bucketed by length. Mini-batches are drawn from these buckets.
\item When the {\em curriculum update frequency} is reached (defined in terms of number of batches), the curriculum's schedule is updated. For example, this may imply that we include more difficult shards in training in the next phase. In cases where the total number of examples in these shards is smaller than the curriculum update frequency, we repeat the previous step until the update frequency has been achieved. 
\item After all available shards are visible to the model, training continues until validation perplexity does not improve for 32 checkpoints. The NMT model has then {\em converged}.

\end{itemize}

%% file: setup.tex
\externaldocument{results.tex}
\externaldocument{methods.tex}
\section{Experiment Setup}
\label{sec:setup}

\paragraph{Data} All experiments were conducted on the German-English parallel dataset from the Multi-target TED Talks Task (MTTT) corpus \cite{duh18multitarget}. The \textit{train} portion consists of about 150k parallel sentences while the \textit{dev} and \textit{test} subsets have about 2k sentences each. All subsets were tokenized and split into subwords using byte pair encoding (BPE) \cite{P16-1162}. The BPE models were trained on the source and target language separately and the number of BPE symbols was set to 30k.

\paragraph{NMT Setup} Our neural machine translation models were trained using Sockeye\footnote{\url{github.com/awslabs/sockeye}} \cite{hieber2017sockeye}. We used 512-dimensional word embeddings and one LSTM layer in both encoder and decoder. We used word-count based batching (4096). Our systems employed the Adam optimizer \cite{DBLP:journals/corr/KingmaB14} with an initial learning rate of either 0.0002 or 0.0008 (see Section~\ref{sec:results}). The \textit{dev} set from the corpus was used as a validation set for early stopping. 

The baseline is an NMT model with the structure and hyperparameters described above  without a curriculum; that is, it has access to the entire training set which is bucketed by length to then create mini-batches.
Training data are split randomly into the same number of shards as the curriculum models (5 here). 

We build the auxiliary model for the use of generating {\em one-best score} for each training sample, with similar but simpler configurations compared to the baseline model, in terms of number of RNN hidden units (200 vs. 512). While the training time for this specific model may cancel out the time saved by curriculum learning in practice, having a high-quality {\em one-best score} provides a useful reference point for our understanding of curriculum learning. 
\marine{clarify how this models differs from baseline; we need to acknowledge somehow that this is not a realistic training procedure}
\xuan{Edited accordingly.}
\marine{Thanks! I'm softening it a bit so it sounds a little more positive.}

\paragraph{Curriculum Learning Setup} The curriculum learning framework as described in Section~\ref{sec:methods} was implemented within Sockeye. Curriculum learning can be enabled as an alternative to default training within Sockeye by specifying a file which contains sentence level scores (difficulty ranking per sentence with respect to any criterion). This implementation leverages the Sockeye sharding feature, which was originally meant for data parallelism. The codebase is publicly available with our experimental settings and tutorials\footnote{\url{https://github.com/kevinduh/sockeye-recipes/tree/master/egs/curriculum}}.

We set the curriculum's update frequency to 1000 batches, which is the same as our checkpoint frequency.

%% file: results.tex
\section{Results}
\label{sec:results}

\begin{figure}[ht]
\centering
\includegraphics[width=\linewidth]{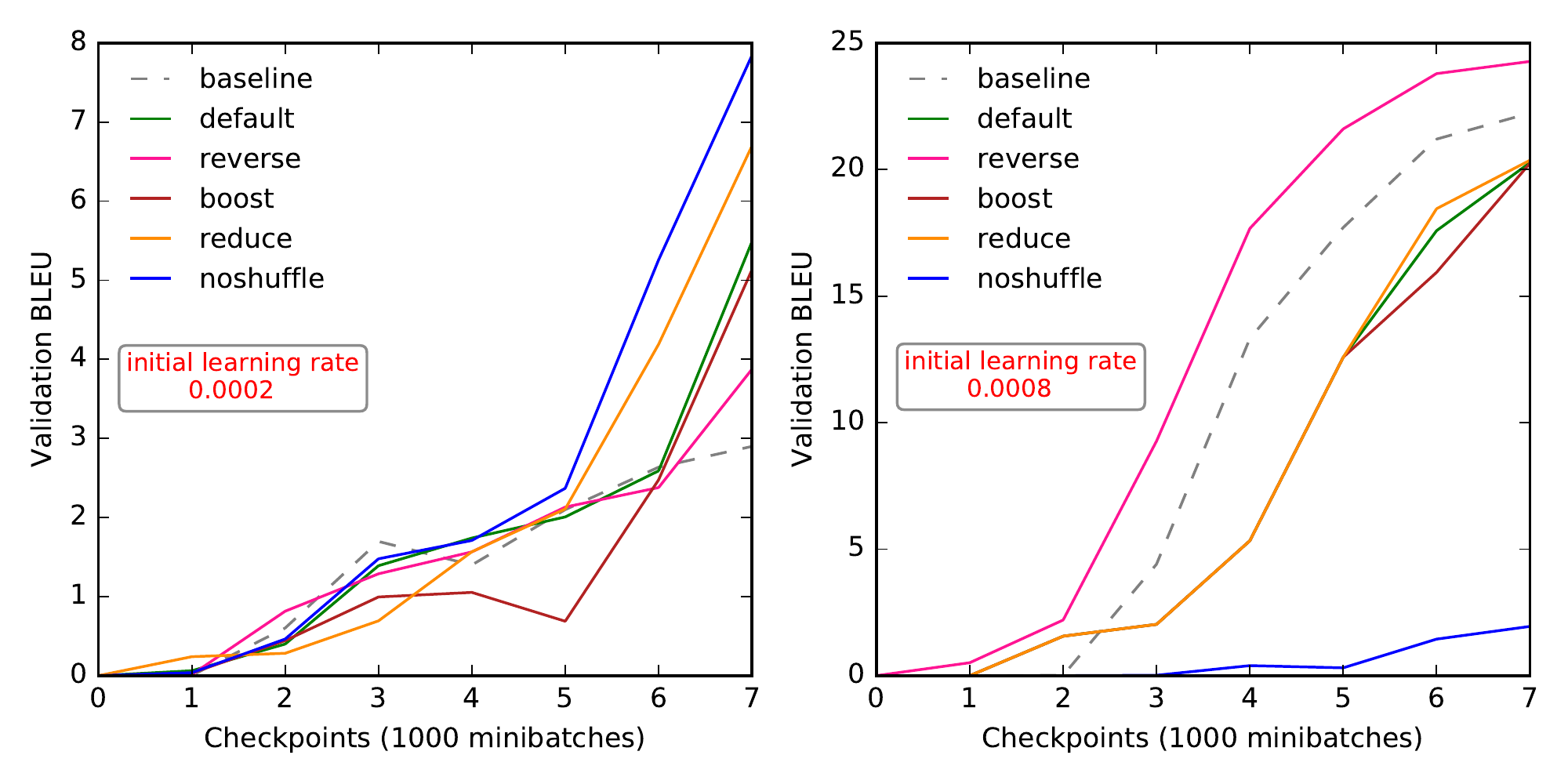}
\caption{Learning curves for the first 7 curriculum updates. The NMT model is trained on data organized by the {\em avg word freq rank (de)} difficulty criterion with different curriculum learning schedules. }
\label{fig:learning_first7}
\end{figure}

We start by examining training behavior during early training stages. Figure~\ref{fig:learning_first7} shows the learning curves (validation BLEU\footnote{BLEU is the standard evaluation for machine translation based on n-gram precision; higher is better \cite{bleu}.} vs checkpoints) for the first 7 checkpoints\footnote{7 is the lowest number of checkpoints required to discriminate between the different schedules.} of curriculum training.  The curriculum is updated at each checkpoint using one of the schedules listed in section~\ref{ssec:curriculumschedule}.
With the smaller learning rate, all curricula improve over baseline validation BLEU at the 7th checkpoint. However, with the higher learning rate, only the {\em reverse} schedule outperforms the baseline.
Similar trends are observed with other difficulty criteria:\footnote{All learning curves available in Supplemental Material} a few curriculum schedules beat the baseline but this outcome is sensitive to the initial learning rate.
 
\begin{table}[ht]
\centering
\small
\begin{tabular}{c|r|r|r}
\hline
{\bf Curr Update} & {\bf Time} & {\bf BLEU} & {\bf BLEU} \\
{\bf Freq} & (thousand batches) & (7) & (best)\\
\hline
1000&108&8.8&28.2\\
2000&100&1.8&28.0\\
3000&71&9.2&28.2\\
4000&56&9.0&27.9\\
5000&108&14.9&28.0\\
6000&67&14.9&28.0\\
\hline
\end{tabular}
\caption{Impact of curriculum update frequency on the model trained on {\em default} schedule with data organized by {\em avg word freq rank (de)}. Training time is quantified as total number of mini-batches the NMT model has processed before convergence. The initial learning rate is set to 0.0002. The last two columns show the decoding performance of the model at 7th and the {\em best checkpoint} --- the checkpoint at which the model got highest BLEU score on val set.}
\label{tab:updatefreq}
\end{table}

\input{tables.tex}
When training until convergence (Tables~\ref{tab:0.0002best}-\ref{tab:0.0008best}),  20 of 100 curriculum strategies successfully converge earlier than the baseline without loss in BLEU. The model trained with the average source word frequency as a difficulty criterion and the {\em reverse} schedule improves training time by 19\% to 30\%.\footnote{These are substantial time savings given that training the baseline took up to 1 day.}
However, the optimal curriculum schedule for other complexity criteria change with the initial learning rate. The model trained with the {\em one-best score} and the {\em boost} schedule converges after processing 19\% fewer mini-batches than the baseline (59,000 vs. 73,000) and yields a comparable BLEU score (28.4 vs. 28.1) with an initial learning rate of 0.002. With a higher initial learning rate, this configuration also speeds up training by 38\% (48,000 vs. 79,000) but at the cost of a 1.65 point degradation in BLEU. The default schedule yields better results with the learning rate of 0.0008 but not 0.0002.

Comparing trends across complexity criteria shows there is no clear benefit to the expensive one-best model score compared to the simpler word frequency criteria. Sentence length is not a useful criterion: it helps convergence time only slightly (74,000 vs. 79,000) and in only one of the ten configurations we run.\marine{I wonder whether we're being hit by a brevity penalty with sentence length curricula?} This is a surprising result at first, given that both sentence length and word frequencies were found to be useful ordering criteria by \citet{ZhangKimCregoSenellart2017}. However, their experiments are not directly comparable. They were limited to a single training epoch and use a different training strategy, which is closest to our {\em noshuffle} schedule. With that schedule, our de-en sentence length curricula also outperform the baseline in early training stages, but the baseline catches up and outperforms by convergence time.
\Gaurav{Compare results to papers which propose length based features and boost/reduce for curriculum learning. State that their results do not hold for our datasets/conditions.}
We also note that the conclusions about the {\em reduce} stated by \citet{ZhangKimCregoSenellart2017} do not hold true for our dataset and curriculum schedules. Specifically, this schedule provides no improvement in training time. (Table~\ref{tab:0.0002best} and~\ref{tab:0.0008best}).

These results highlight the benefits of viewing curriculum learning broadly, and of curriculum strategies beyond the initial ``easy samples first'' hypothesis. Interestingly, the {\em default} and {\em reverse} schedules can yield close performance, and forcing data shards to be explored in order ({\em noshuffle}) does not  improve over the {\em default} sampling schedule.

Table~\ref{tab:updatefreq} further illustrates how curriculum training in NMT is sensitive to hyperparameters. We change the curriculum update frequency (mini-batches) and notice that while the validation set BLEU ramps up quickly as the number of mini-batches is increased between curriculum updates, the convergence time shows no clear trend and the validation BLEU at convergence is the same. 

\Gaurav{Overall trends need to restated; 1. Simple length based features do as well as 1-best etc. 2. No clear winners emerge. 3. LR and other hyperparameters matter}
To sum up, our extensive experiments show that curriculum learning can improve convergence speed, but the choice of difficulty criteria is key: vocabulary frequency performs as well as the more expensive one-best score, and sentence length does not help beyond early training stages. No single curriculum schedule consistently outperforms the others, and results are sensitive to other hyperparameters such as initial learning rate and curriculum update frequency.

\section{Conclusion}

We investigated whether curriculum learning is effective in speeding up the training of complex neural network models such as those used in neural machine translation (NMT) on a German-English TED translation task. NMT is a good test case for curriculum learning as training is prohibitively slow and much patience is required to reach good performance. While the impact on other language pairs and datasets remains to be studied, we contribute an extensive exploration of curriculum design in controlled settings.
We adopt a probabilistic view of curriculum learning, implemented on top of a state-of-the-art NMT toolkit, in order to enable a flexible evaluation of the impact of various curricula design.
Our contribution is an extensive exploration of various ways to design the curriculum, both in terms of the difficulty criteria and the curriculum schedule.
Our conclusions can be interpreted both positively and negatively:  Our results demonstrate curriculum learning can be an effective method for training expensive models like those in NMT, as 20 of the 100 curricula tried improved convergence speed at no loss in BLEU, and that ``easy to hard'' is not the only useful sample ordering strategy. However, careful design of the specific curriculum hyperparameters is important in practice.

%% file: tables.tex
\begin{table*}[ht]
\centering
\small
\begin{tabular}{ l| rrrrr | c  c  c c c}
\hline
& \multicolumn{5}{|c|}{{\bf Training Time} (thousand batches)} & \multicolumn{5}{|c}{{\bf Test BLEU} (best)} \\
\hline
baseline & \multicolumn{5}{|c|}{73} & \multicolumn{5}{|c}{28.1}\\
\hline
& {\em default} & {\em reverse} & {\em boost} & {\em reduce} & {\em noshuffle} & {\em default} & {\em reverse} & {\em boost} & {\em reduce} & {\em noshuffle} \\
\hline
{\em one-best score} & 56&80&{\bf 59}&64&92&
27.0&27.9&{\bf 28.4}&27.3	&27.4 \\
\hline
{\em max wd freq(de)} &
57&88&89&82&77&25.2&26.1&27.4&27.2&28.1\\
{\em max wd freq(en)} & 
{\bf 63}&77&75& 64&98&
{\bf 27.6}&25.3&27.5&26.9&27.6\\
{\em max wd freq(deen)} &
{\bf 56}&61&{\bf 62}&{\bf 59}&{\bf 62}&
{\bf 28.1}&27.5&{\bf 27.8}&{\bf 27.7}&{\bf 28.5}\\
\hline
{\em ave wd freq(de)} &
{\bf 72}&{\bf 69}&57&73&108&
{\bf 28.2}&{\bf 28.5}&27.3&26.5&28.2\\
{\em ave wd freq(en)} &
84&66&61&61&{\bf 64}&
27.8&25.4&27.4&25.8&{\bf 27.9}\\
{\em ave wd freq(deen)} &
62& 57&84&85&{\bf 67}&
27.3&27.4&28.3&26.9&{\bf 28.2}\\
\hline
{\em sent len(de)} &
78&118&67&56&83&
26.6&28.1&27.2&26.4&27.6\\
{\em sent len(en)} &
151&59&67&125&196&27.6&25.1&25.6&27.1&27.7\\
{\em sent len(deen)} &
113&189&79&68&195&27.0&26.3&26.3&23.9&27.7\\
\hline
\end{tabular}
\caption{Performance of curriculum learning strategies with initial learning rate 0.0002. Training time is defined as in Table \ref{tab:updatefreq}. Bold numbers indicate models that win on training time with comparable (difference is less or equal to 0.5) or better BLEU compared to the baseline.}
\label{tab:0.0002best}
\end{table*}

\begin{table*}[ht]
\centering
\small
\begin{tabular}{ l| rrrrr | c c c c c}
\hline
& \multicolumn{5}{|c|}{{\bf Training Time} (thousand batches)} & \multicolumn{5}{|c}{{\bf Test BLEU} (best)} \\
\hline
baseline & \multicolumn{5}{|c|}{79} & \multicolumn{5}{|c}{29.95}\\
\hline
& {\em default} & {\em reverse} & {\em boost} & {\em reduce} & {\em noshuffle} & {\em default} & {\em reverse} & {\em boost} & {\em reduce} & {\em noshuffle} \\
\hline
{\em one-best score} & {\bf 59}&{\bf 69}&48&92&112&
{\bf 30.1}&{\bf 29.9}	&28.3&28.9&30.4\\
\hline
{\em max wd freq (de)} & 85	&103	&{\bf 69}&	118	&{\bf 43}&	
25.9	&29.6&	{\bf 30.7}&	25.8	&{\bf 29.6}\\

{\em max wd freq (en)} & 148&	80	&166&	49	&158&	
27.0	&29.6	&28.4	&{\bf 29.5}	&29.9\\

{\em max wd freq (deen)} &
84&	{\bf 61}	& {\bf 75}	& 67	&93&	
29.5	&{\bf 31.5}&	{\bf 31.1}	&27.9	&27.2\\
\hline
{\em ave wd freq (de)} &
79&	{\bf 51}&73&	88	&58&	
27.3	&{\bf 30}	&27.6	&27.1	&21.3\\

{\em ave wd freq (en)} &
{\bf 72}	& 71	&146	&61	&74	&
{\bf 29.9} &	28.4	&23.3	&25.2	&29.4\\

{\em ave wd freq (deen)} &
81&	47&	54	& 58	&71	&
29.9	&28.4&	28.5	&28.3& 29.3\\
\hline
{\em sent length (de)} &
49&	126	&88	&85&	{\bf 74}	&
27.0	&30.3	&29.3&	27.8&	{\bf 31.0}\\

{\em sent length (en)} &
101	&52&70	&49	&114&	
29.0&	27.6&	24.2&	26.9	&30.2\\

{\em sent length (deen)} &
155	&148	&170&	95	&86	&
29.4	&30.7&30.5&	29.6	&29.5\\
\hline
\end{tabular}
\caption{Performance of curriculum learning strategies with initial learning rate 0.0008.}
\label{tab:0.0008best}
\end{table*}

%% file: sm.tex
\section{Supplementary Material}
\begin{figure*}[hbt!]
	\centering
	\includegraphics[width=\linewidth]{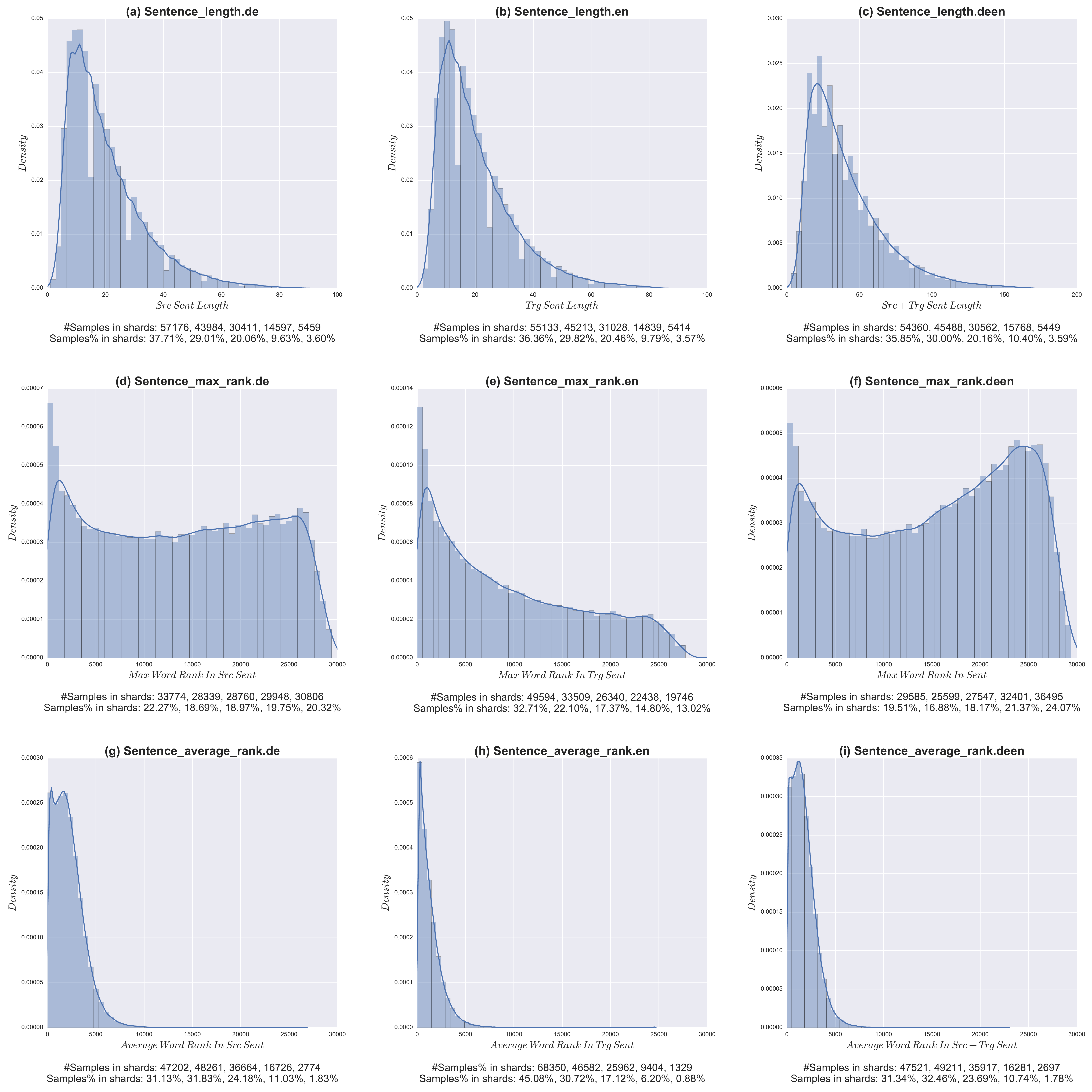}
	\caption{Statistics on GE-EN TED Talks training set (151,627 samples in total) scored by different difficulty criteria. We split the training data into 5 shards. Bucketing results using Jenks Natural Breaks classification algorithm are shown below each subplot, starting from easiest shard to harder shards.}
    \label{fig:datastat}
\end{figure*}

\begin{figure*}
	\includegraphics[width=\linewidth]{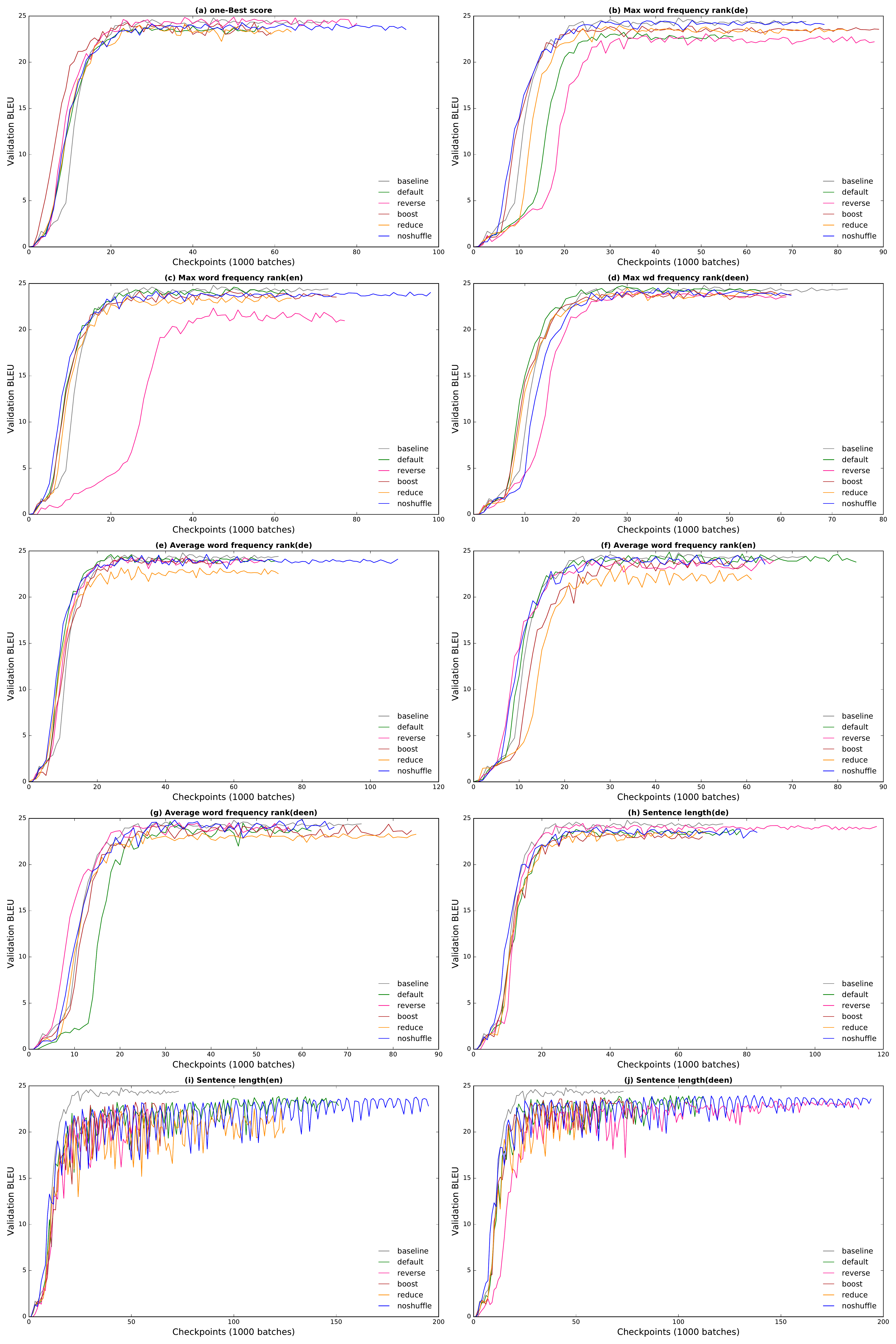}
    \label{fig:score2}
    \caption{Validation BLEU curves with initial learning rate 0.0002 for different sample ranking criteria.}
\end{figure*}

\begin{figure*}
	\includegraphics[width=\linewidth]{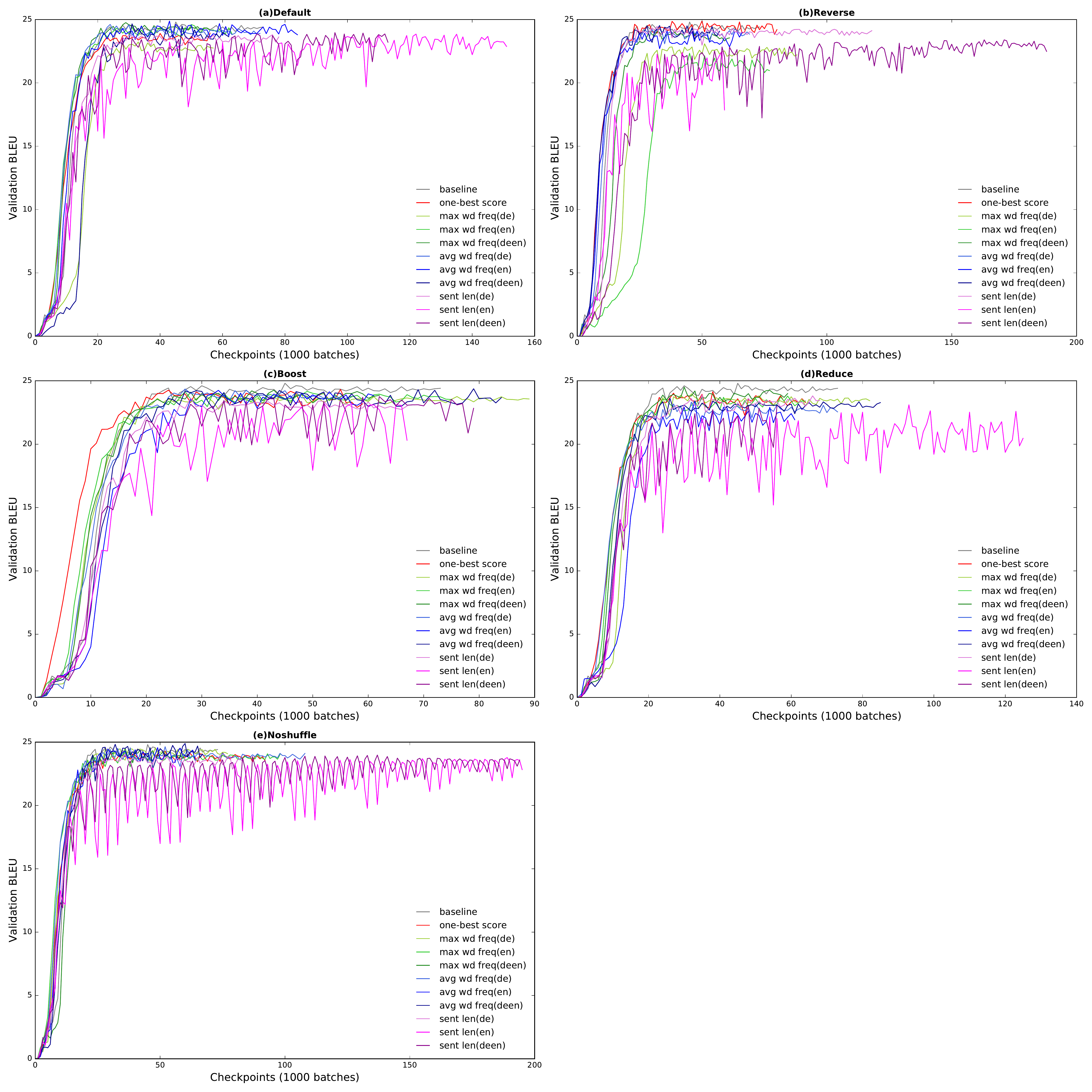}
    \label{fig:schedule2}
    \caption{Validation BLEU curves with initial learning rate 0.0002 for different curriculum schedules.}

\end{figure*}

\begin{figure*}
	\includegraphics[width=\linewidth]{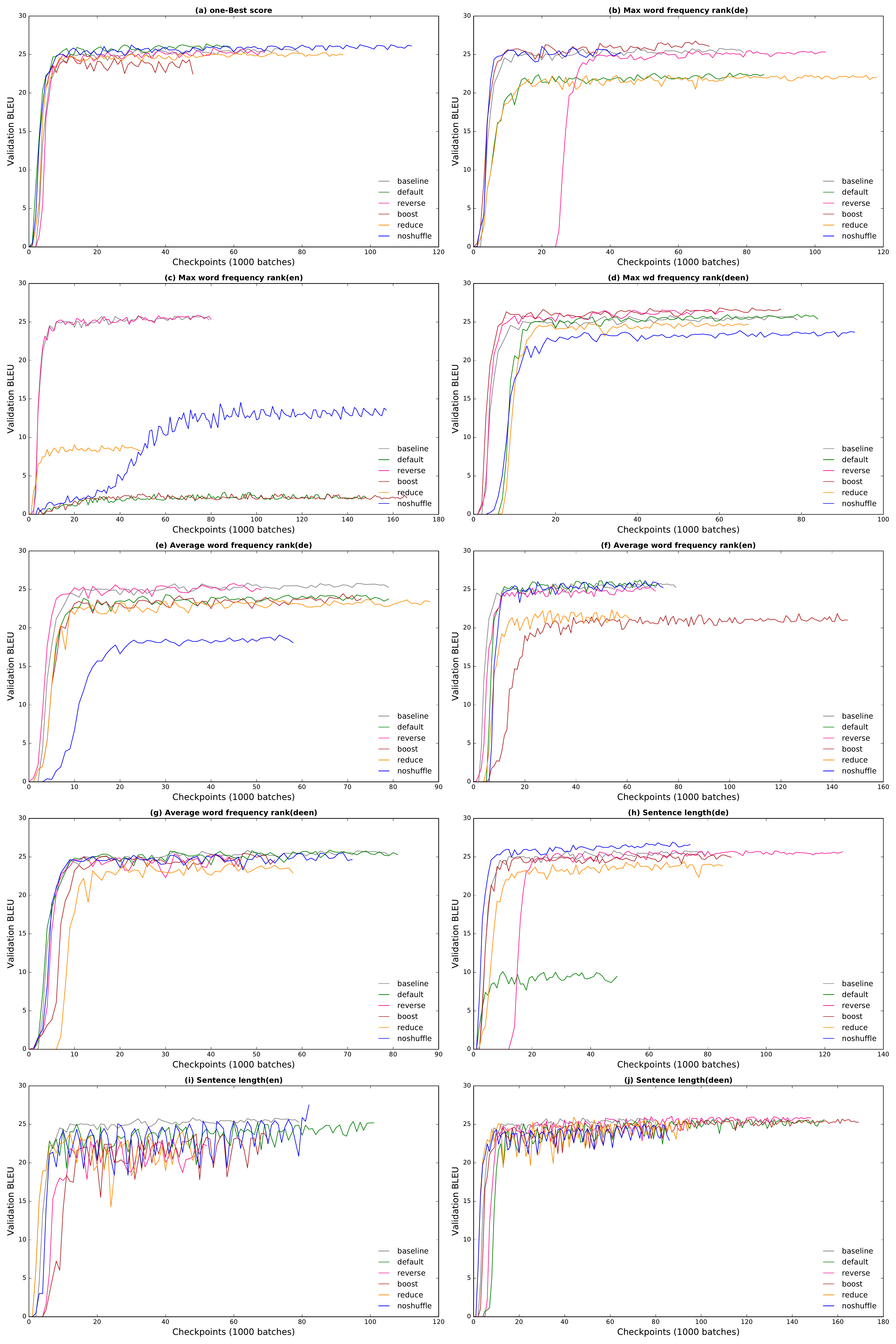}
    \label{fig:score8}
    \caption{Validation BLEU curves with initial learning rate 0.0008 for different sample ranking criteria.}
\end{figure*}

\begin{figure*}
	\includegraphics[width=\linewidth]{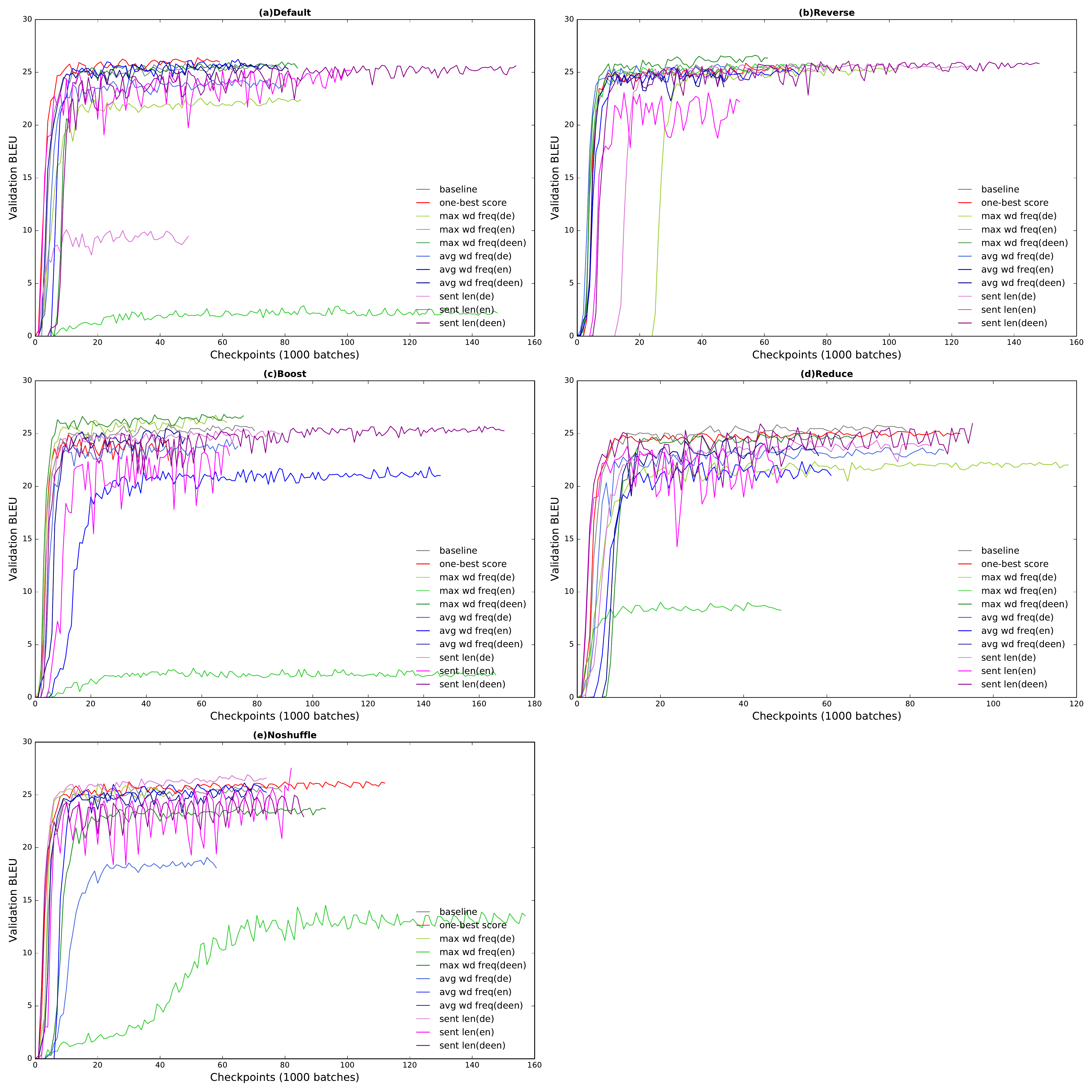}
    \label{fig:schedule8}
    \caption{Validation BLEU curves with initial learning rate 0.0008 for different curriculum schedules.}
\end{figure*}
\input{table}

%% file: table.tex
\begin{table*}[!h]
\small
\centering
\begin{tabular}{c| c c c c c}
\hline 
baseline & \multicolumn{5}{|c}{2.84} \\
\hline
& {\em default} & {\em reverse} & {\em boost} & {\em reduce} & {\em noshuffle} \\
\hline 
{\em one-best score} & {\bf 7.1}&{\bf 9.2}&{\bf 14.7}&{\bf 7.8}&{\bf 7.8}\\
\hline
{\em max wd freq(de)} & 2.0&1.7&{\bf 4.6}&1.6&{\bf 7.8}\\
{\em max wd freq(en)} & {\bf 7.9}&0.8&{\bf 8.1}&{\bf 5}&{\bf 10.8}\\
{\em max wd freq(deen)} & {\bf 4.3}&2.5&{\bf 4.2}&{\bf 2.9}&2.2\\
\hline
{\em avg wd freq(de)} & {\bf 6.5}&{\bf 4.1}&{\bf 5.5}&{\bf 7.5}&{\bf 8.8}\\
{\em avg wd freq(en)} & 2.7&{\bf 6.8}&2.2&2.6&{\bf 5.8}\\
{\em avg wd freq(deen)} & 1.6&{\bf 8.7}&{\bf 2.9}&1.7&{\bf 3.7}\\
\hline 
{\em sent len(de)} & 2.4&{\bf 3.0}&{\bf 2.9}&1.6&{\bf 4.6}\\
{\em sent len(en)} & {\bf 3.3}&{\bf 3.3}&2.5&2.1&{\bf 5.1}\\
{\em sent len(deen)} & 2.0&2.2&2.0&2.0&{\bf 4.3}\\
\hline
\end{tabular}
\caption{Decoding performance of different curriculum learning models at the 7th checkpoint with initial learning rate 0.0002.}
\label{tab:0.0002cp7}
\end{table*}

\begin{table*}
\centering
\begin{tabular}{c|c c c c c}
\hline 
baseline & \multicolumn{5}{|c}{25.1}\\
\hline
& {\em default} & {\em reverse} & {\em boost} & {\em reduce} & {\em noshuffle}\\
\hline 
{\em one-best score} &{\bf 28.6}&	3.7	&{\bf 26.5}	&{\bf 26.3}&	{\bf 27.8}\\
\hline
{\em max wd freq(de)} &18.8	& 0.0 &	{\bf 27.9}&	18.6&	{\bf 29.1}\\
{\em max wd freq(en)} &0.4	&{\bf 26.7}&	0.0	&8.5	&0.7\\
{\em max wd freq(deen))} &2.0		&{\bf 28.3}&{\bf 28.3}&0.2&5.2\\
\hline
{\em avg wd freq(de)} &24.1	&{\bf 28.8}	&23.3&	23.1&	2.3\\
{\em avg wd freq(en)} &18.1&21.5	&1.9	&9.7	&4.9\\
{\em avg wd freq(deen)} & {\bf 25.2}	&{\bf 26.4}&	18.5&	2.0&	{\bf 26.4}\\
\hline
{\em sent len(de)} &9.9		&0.0&24.3&	17.6	&{\bf 30.0}\\
{\em sen len(en)} & {\bf 26.6}	&18.6	&5.3	&{\bf 26.1}	&24.2\\
{\em sent len(deen)} & 1.1	&14.4	&24.1	&{\bf 26.0}&	24.2\\
\hline
\end{tabular}
\caption{Decoding performance of different curriculum learning models at the 7th checkpoint with initial learning rate 0.0008.}
\label{tab:0.0008cp7}
\end{table*}